\documentclass{article} 
\usepackage{iclr2026_conference,times}

\usepackage{amsmath,amsfonts,bm}

\def\eqref#1{equation~\ref{#1}}

\def\1{\bm{1}}

\DeclareMathAlphabet{\mathsfit}{\encodingdefault}{\sfdefault}{m}{sl}
\SetMathAlphabet{\mathsfit}{bold}{\encodingdefault}{\sfdefault}{bx}{n}

\usepackage{hyperref}
\usepackage{url}

\usepackage[utf8]{inputenc} 
\usepackage[T1]{fontenc}    
\usepackage{hyperref}       
\usepackage{url}            
\usepackage{booktabs}       
\usepackage{amsfonts}       
\usepackage{nicefrac}       
\usepackage{microtype}      
\usepackage{xcolor}         
\usepackage{amsmath} 
\usepackage{listings}
\usepackage{tabularx}
\usepackage{tcolorbox}
\usepackage{multirow}
\usepackage[table]{xcolor} 
\usepackage{marvosym}

\usepackage{bbding}
\usepackage{multicol}
\usepackage{pifont}

\usepackage{graphicx}
\usepackage{wrapfig} 
\usepackage{lipsum}
\tcbuselibrary{breakable}
\usepackage{listings}

\title{CogOmniControl: Reasoning-Driven Controllable Video Generation via Creative Intent Cognition}

\author{
\centerline{\textbf{Hongji Yang\textsuperscript{1,*},\quad
Songlian Li\textsuperscript{2,*},\quad
Yucheng Zhou\textsuperscript{1},\quad
Xiaotong Zhao\textsuperscript{2}}} \\[2pt]
\centerline{\textbf{Alan Zhao\textsuperscript{2},\quad
Chengzhong Xu\textsuperscript{1},\quad
Jianbing Shen\textsuperscript{1,\Letter}}} \\[4pt]
\centerline{\textsuperscript{1}SKL-IOTSC, CIS, University of Macau \quad
\textsuperscript{2}Online-Video BU, Tencent} \\[2pt]
\centerline{\textsuperscript{*}Equal contribution. \quad
\textsuperscript{\Letter}Corresponding author.}
}

\iclrfinalcopy 
\begin{document}

\maketitle

\begin{abstract}

Recent diffusion models achieve strong photorealism and fluency in video generation, yet remain fragile under abstract, sparse or complex conditions, leading to poor performance in professional production workflows such as storyboard sketches and clay render conditions. Existing video generation models, either inject conditions through adapters or couple a generic vision-language model (VLM) within a diffusion backbone, leaving a capability gap and failing to produce the videos that align with the user's creative intent. We present \textbf{CogOmniControl}, a reasoning-driven framework that factorizes controllable video generation into creative intent cognition and generation. Specifically, we train a specialized CogVLM using authentic anime production data. Compared to generic VLMs, it generates more professional and clear outputs, accurately cognizing user creative intent from sparse and abstract conditions and tuning these cues into dense reasoning output. Besides, CogOmniDiT unifies the controls from various conditions through in-context generation and is aligned to the CogVLM reasoning outputs via reinforcement learning. Furthermore, leveraging CogVLM's robust capability in guiding video generation, we release its potential in planning specific evaluators and enable a Best-of-N selection for the generated videos. This integration transforms the entire framework into a closed-loop ``harness-like'' architecture. We further introduce CogReasonBench and CogControlBench, built from professional workflows data that carry genuine creative intent rather than simulated ones. Experiments on two benchmarks show that CogOmniControl surpassed the existing open-source models.
The project website: \url{ https://um-lab.github.io/CogOmniControl/}.

\end{abstract}

\vspace{-2mm}
\section{Introduction}

\begin{figure}
    \vspace{-2mm}
    \centering
    \includegraphics[width=1.0\linewidth]{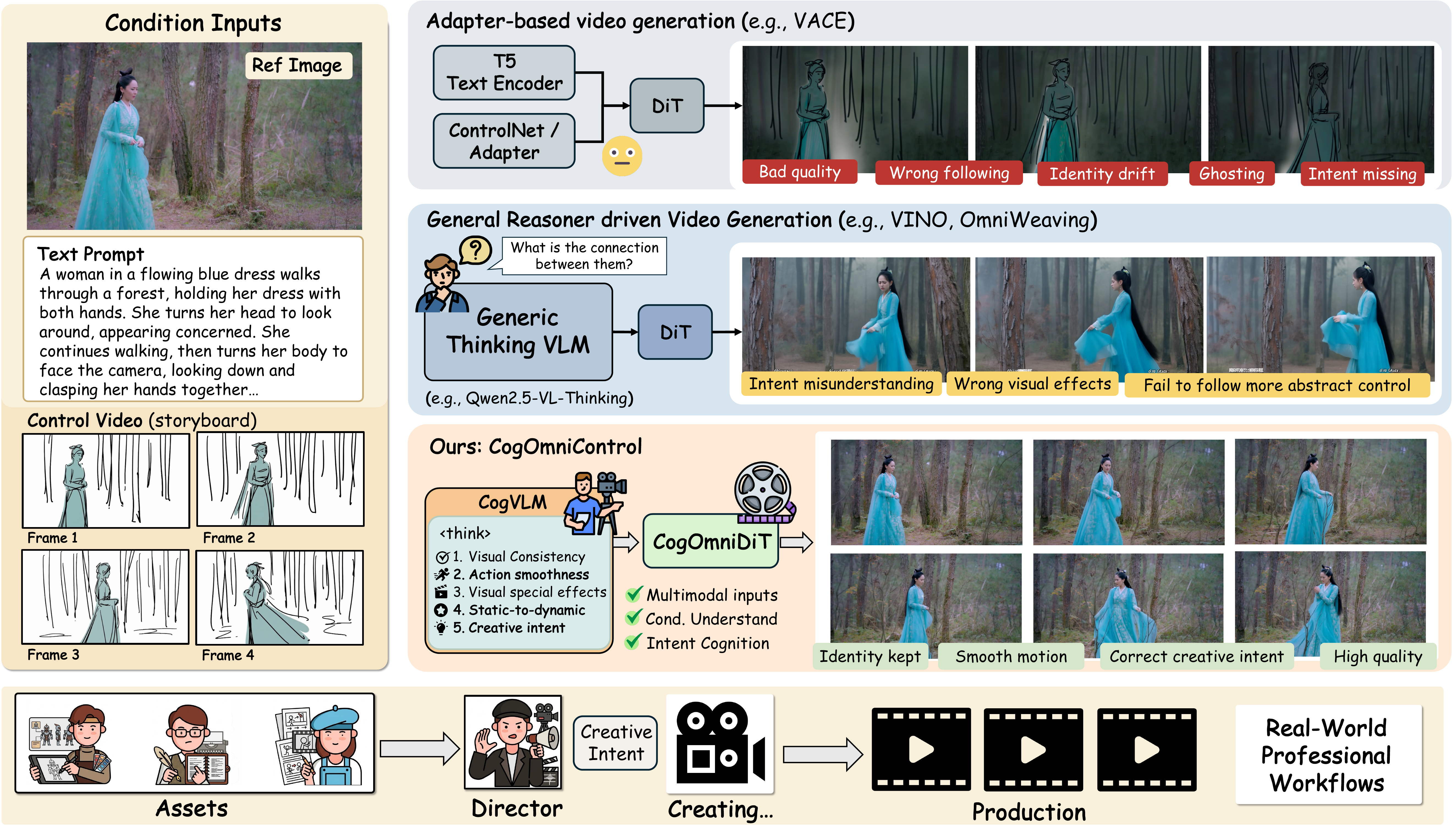}
    \vspace{-8mm}
    \caption{The motivation of our CogOmniControl. The adapter-based methods and video generation models with generic VLM fail to generate the final video from the given condition. }
    \label{fig:motivation}
    \vspace{-6mm}
\end{figure}

Recent advances in diffusion-based video generative models~\citep{hong2022cogvideo, yang2024cogvideox, hacohen2024ltx, wan2025} have pushed text-to-video generation to a level of photorealism and motion fluency. Current research~\citep{jiang2025vace, pan2026omniweaving} is moving toward omni-level controllable generation, pursuing a single system to support multimodal inputs, professional intent conditions and abstract constraints. Inspired by the powerful multimodal understanding capabilities of VLMs, these frameworks~\citep{tan2025omni, yang2026omni,pan2026omniweaving} attempt to employ VLMs to identify and correlate different condition inputs and then cognize the creative intents to infer coherent control signals. 
However, video generation still faces the key challenges: 
\ding{172}~\textbf{Cognitive Gap}: When confronted with complex or even conflicting multimodal control signals in professional workflows, current VLMs struggle to fully comprehend the underlying creative intent. Consequently, they fail to formulate reasonable generation plans grounded in domain-specific creative knowledge. 
\ding{173}~\textbf{Alignment Gap}: It remains an open question whether the outputs of VLMs under abstract conditions are properly aligned with the generated videos. Besides, the adoption of reasoning output from generic VLM also brings additional noise~\citep{yang2026omni,chen2026vino} for the generation.
As shown in Fig.~\ref{fig:motivation}, it remains challenging for controllable video generation models to understand abstract conditions, infer creative intent, and then generate correct video outputs. 

To bridge this gap between the abstract condition and creative intent, we present CogOmniControl, which includes the CogVLM to cognize the creative intent and CogOmniDiT to transform the intent into video output. 
To enable VLMs to understand abstract conditions and creative intent for more efficient reasoning, we employ a combination of Supervised Fine-Tuning (SFT) and Reinforcement Fine-Tuning (RFT). This process transforms a generic VLM into a specialized CogVLM, equipped with deeper controllable video generation knowledge to more effectively drive video generation models. 
By incorporating high-level features from CogVLM and conditional inputs, CogOmniDiT achieves more robust controllable generation with abstract and sparse conditions. 
Unlike previous approaches that simulated user intent from existing videos, our dataset was collected from real-world professional workflows, including the storyboard, clay rending video, and their corresponding video, which represent genuine creative intent from initial sketches to final production. 
Drawing on LLM harness engineering~\citep{eval-harness, lee2026meta, lin2026agentic}, CogVLM goes beyond specifying the generation for the DiT, it can also identify the required evaluators derived from its reasoning through the conditions. This enables the model to pick suitable evaluators for optional Best-of-N selection, establishing a fully integrated closed-loop pipeline in video generation.
We also define a suite of tools as evaluators, including both VLMs and specialized pre-trained models in the framework.
To further evaluate the understanding of abstract conditions and the quality of video generation from both VLM and video generation models, we introduce two benchmarks, CogReasonBench and CogControlBench, to validate our proposed method.
Experimental results on two benchmarks demonstrate that our model outperforms existing open-source models.
Our contribution can be summarized as follows:
\begin{itemize}
    \item We present CogOmniControl, a reasoning-driven framework for controllable video generation. By leveraging professional reasoning to bridge the gap between pixel-level priors and high-level intent, our framework ensures structural integrity and creative intent alignment, particularly in sparse and abstract controllable generation scenarios.
    \item We propose CogVLM and CogOmniDiT. CogVLM understands abstract and sparse conditions, infers the creative intent, and translates multimodal cues into dense logical outputs. CogOmniDiT integrates diverse control signals with the high-level semantic features from CogVLM, faithfully synthesizing videos aligned with the inferred intent.
    \item We further extend CogOmniControl into a closed-loop Reasoning-Generation-Verification system through an evaluator harness emitted by CogVLM. In a single forward pass, CogVLM produces a solution as well as the evaluator, which scores the candidates in Best-of-N selection.
    \item To evaluate the conditions understanding and abstract reasoning of VLM and instruction following of controllable video generation, we construct two new benchmarks, CogReasonBench and CogControlBench, for CogVLM and CogOmniControl, respectively. These benchmarks are collected from human-drawn storyboards or clay render videos during real-world professional animation productions.
\end{itemize}

\section{Related Work}
\noindent\textbf{Video Generation.}
With the rapid development of image~\citep{rombach2022high,podell2023sdxl,peebles2023scalable, flux2024} and video~\citep{hong2022cogvideo,yang2024cogvideox,hacohen2024ltx,kong2024hunyuanvideo,wan2025} generative models, diffusion models have been proven to produce high-fidelity visual content and are widely applied in diverse domains, including artistic creation, animation production, visual special effects and game development~\citep{brooks2024video,Midjourney}. 
To faithfully realize specific creative intentions, conditional guidance has evolved from abstract natural language to diverse explicit constraints for precise control. Early breakthroughs introduce additional adapter~\citep{zhang2023adding, ye2023ip, li2025controlnet, yang2025dc, guo2023animatediff, zhao2023controlvideo, jiang2025vace,guo2024sparsectrl,lin2024ctrl, liu2025sketchvideo} to support condition injection without compromising the original generative quality. 
However, these adapter-based paradigms often exhibit limited flexibility in handling diverse conditions, particularly in those that are non-pixel-aligned or serve merely as visual references. To achieve omni-level, OmniGen~\citep{xiao2025omnigen} and OmniGen2~\citep{wu2025omnigen2} integrated autoregressive transformers with diffusion to realize a unified generation. OmniControl~\citep{tan2025ominicontrol} and UNO~\citep{wu2025less} introduced in-content visual generation. 

The omni-level generation has also been extended into the video domain, the emergence of proprietary models, such as Seedance2.0~\citep{seedance2026seedance}, Kling-O1~\citep{team2025kling}, Sora2~\citep{Sora2}, Vidu~\citep{Vidu}, Veo3~\citep{veo3}, has established a transformative vision for omni-level video generation. However, current open-source models still fail to realize robust unified video generation. VACE~\citep{jiang2025vace}, UniVideo~\citep{wei2025univideo}, and VINO~\citep{chen2026vino} attempted to achieve omni-level generation by integrating various basic tasks, they often lack a deep understanding across diverse conditions. In contrast, OmniWeaving~\citep{pan2026omniweaving} successfully incorporated the abstract reasoning of VLM into the video diffusion model to execute complex multimodal compositional tasks. However, the reasoning processes of its LLM components have not yet undergone professional evaluation or systematic benchmarking on creative intentions, leaving the model without sufficient guidance when tackling more challenging tasks. 

\noindent\textbf{Reinforcement Learning for Visual Generation.}
Inspired by the success of LLM fine-tuning using RL from human feedback, RL for visual generation is gaining momentum. For example, DDPO~\citep{black2024training}, Diffusion-DPO~\citep{wallace2024diffusion} and DPOK~\citep{fan2023dpok} introduced Direct Preference Optimization~\citep{rafailov2023direct} into T2I Diffusion to align with human preference.
Motivated by DeepSeep-R1~\citep{guo2025deepseek} using GRPO~\citep{shao2024deepseekmath} to provide more dense rewards through computing relative rewards in a sample group, Flow-GRPO~\citep{liu2025flow} and DanceGRPO~\citep{xue2025dancegrpo} extended this paradigm into flow-matching models~\citep{liu2022flow} by transforming the deterministic ODE formulation into a stochastic SDE, thereby enabling effective online exploration and policy alignment.
Beyond this, several GRPO-based studies~\citep{wang2025pref, li2025mixgrpo, he2025tempflow, yang2025hicogen} have focused on refining reward design to enhance performance in visual generation.


\section{Method}

\subsection{CogOmniControl Framework}
\begin{figure}
    \centering
    \includegraphics[width=1.0\linewidth]{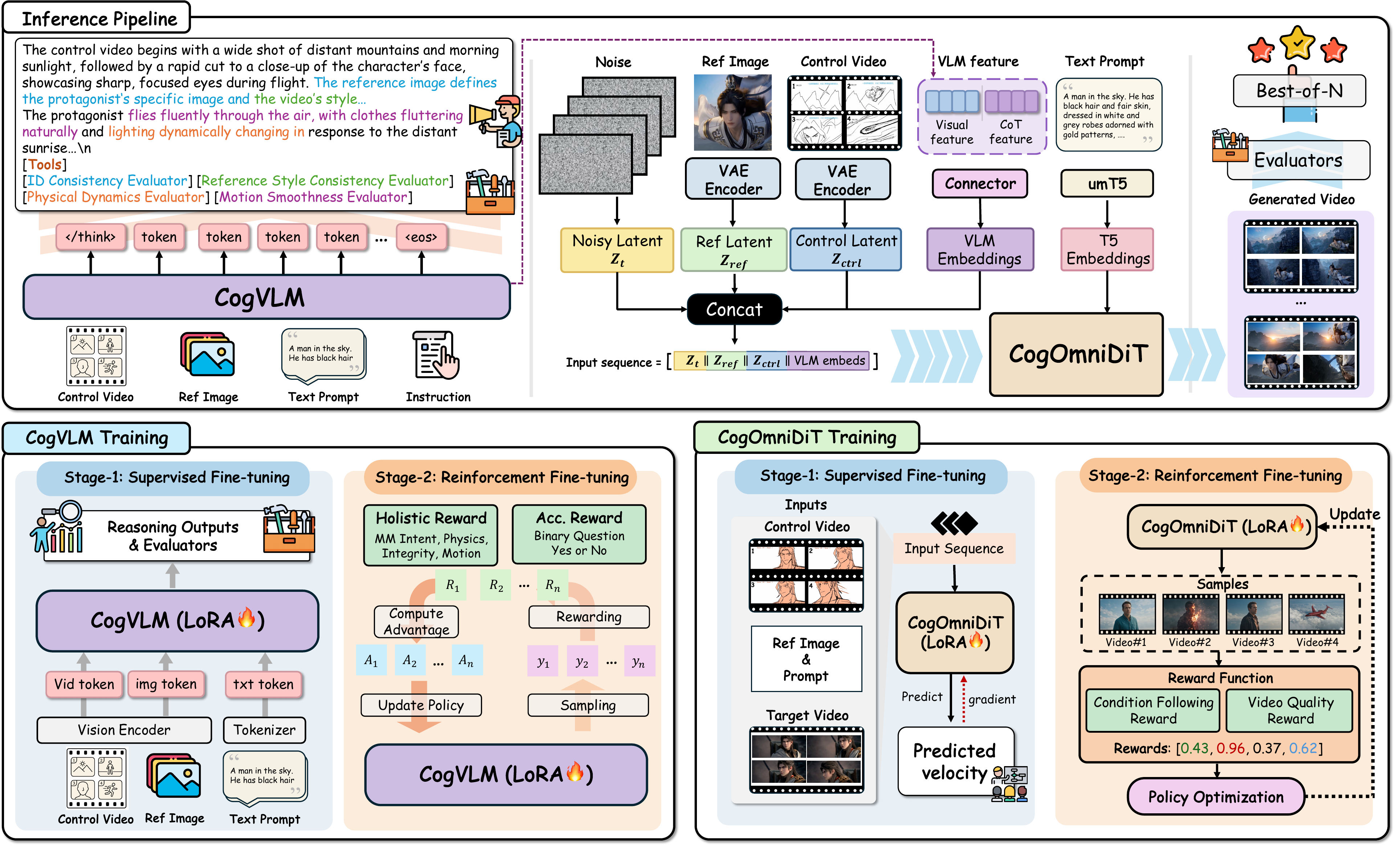}
    \vspace{-4mm}
    \caption{The overall framework of the proposed CogOmniControl. During inference, CogVLM outputs reasoning results based on the given conditions, along with optional evaluator tools. Subsequently, the features from the last layer of CogVLM are concatenated with other latents as the inputs of CogOmniDiT to generate the final result. This process can be repeated multiple times, employing Best-of-N filtering using the evaluator selected by CogVLM. The bottom-left and bottom-right sections illustrate the training processes for CogVLM and CogOmniDiT, respectively.}
    \label{fig:framework}
    \vspace{-4mm}
\end{figure}

In this section, we present the overall framework of CogOmniControl, a robust pipeline that accommodates diverse types of control conditions (e.g., pose, depth, lineart, storyboard sketch, clay render) to facilitate high-quality controllable video generation. 
As illustrated in Fig~\ref{fig:framework}, the proposed method consists of two key modules, CogVLM for reasoning and CogOmniDiT for generation. 

The input condition set $\mathcal{C}$ we define is formulated as a multimodal tuple comprising Control Video $V_{ctrl}$, Reference Image $I_{ref}$ and Textual Description $T_{desc}$, which can be formatted as:
\begin{equation}
    \mathcal{C} = \{V_{ctrl}, I_{ref}, T_{desc}\},
\end{equation}
Control video provides temporal and spatial cues (e.g., trajectories and layouts), the reference image offers visual appearance or spatial references, and the textual description provides global semantic guidance for the entire generation process. 

The core idea of CogOmniControl is to integrate the reasoning of VLM into the controllable generation model.
We formalize the generation process as a conditional mapping $\mathcal{F}$: $ \mathcal{V}\leftarrow{\{V_{ctrl}, I_{ref}, T_{desc}\}}$. Then the whole generation process of CogOmniControl can be formatted as:
\begin{equation}
    P(\mathcal{V}~|~\mathcal{C}) = \underbrace{P(\mathcal{V}~|~\mathcal{{R}}, \mathcal{C} ) }_{\text{Generation}}\cdot  \underbrace{P( \mathcal{{R}}~|~V_{ctrl}, I_{ref}, T_{desc} )}_{\text{Reasoning}},
\end{equation}
\subsection{CogVLM: Cognizing Creative Intent from Multimodal Conditions}
Given a variety of conditions, we observe that they play distinct roles during the creative process. For example, some conditions (i.e., reference images) provide visual information, pose and depth conditions impose strict spatial layouts, and some conditions (i.e., storyboard) may carry additional creative intent. 
However, previous controllable video generation models often treat input conditions as direct pixel-level constraints and fail to align with the creative intent, particularly when conditions exhibit significant conflicts or semantic discrepancies.
Besides, video generative models primarily lack a deep understanding of the diverse input conditions and the underlying correlations between them, making it difficult to coordinate the final generation.

Therefore, we propose CogVLM to perform visual reasoning on how to generate the final video that aligns with the creative intent from different conditions. CogVLM plays the role of the professional director, which ingests multi-modal drafts to formulate explicit production schemes. Specifically, we prompt the VLM to interpret the given conditions and then identify the corresponding cross-modal entities. By reasoning through conflicting constraints and extrapolating implicit details. For example, given `\textit{raining}' in the text and `\textit{standing water}' in the reference image, the VLM can infer emergent visual features like `\textit{rippling effects on the water's surface}' and then generate a dense response.


\noindent\textbf{Training.}
To empower CogVLM with professional-grade insight, we employ a two-stage training strategy, SFT and RFT. For RFT, we design a Holistic Reward and Fact Verification Reward based on LLM-as-a-Judge~\citep{chen2024mllm} to optimize the fine-tuned model. 

The holistic reward function $\text{R}_{holistic}$ is to assess the qualitative alignment of the reasoning output $\mathcal{R}$ with respect to the input conditions $\mathcal{C}$:
\begin{equation}
    \text{R}_{holistic} = \sum_{k\in \mathcal{K}} w_k \cdot \text{VLM}_{k}(\mathcal{R}, \mathcal{C}),
\end{equation}
where $\mathcal{K} = \{intent,phys,info,dyn\}$ represents the four critical dimensions: Creative Intent, Physical Plausibility, Information Integrity, and Motion description.
The function $\text{VLM}_{k}(\cdot)$ denotes the normalized score assigned by the judge model specifically for dimension $k$, weighted by $w_k$.

To ensure the reasoning is grounded in factual accuracy and avoid hallucinations, we implement the Accuracy Reward function $\text{R}_{acc}$. For each condition set $\mathcal{C}$, the teacher model is asked to return $N$ binary questions $\{q_1, q_2, \dots q_N\}$. Then, the judge model verifies whether the reasoning output $\mathcal{R}$ satisfies these atomic facts $q_i$:
\begin{equation}
    \text{R}_{acc} = \frac{1}{N}\sum_{i=1}^N \text{VLM}(\mathcal{R}, q_i),
\end{equation}
This reward mechanism transforms subjective narrative evaluation into a verifiable accuracy metric.

\subsection{CogOmniDiT: Unified Video Diffusion Transformer }
To enable different condition inputs, we present CogOmniDiT, where heterogeneous conditions and noisy latents are processed within a unified sequence. Leveraging the powerful in-context learning~\citep{zhou2024visual,zhou2026multimodal} of the transformer backbone, the noisy latent and various conditions can model themselves and others within the self-attention. This ensures the conditions are effectively injected into the latent, facilitating precise controllable video generation.
\begin{equation}
    \text{Input Sequence} = \text{Concat}(Z_{t}, Z_{ref}, Z_{ctrl}, Emb_{\text{VLM}}),
\end{equation}
where $Z_{t}$, $Z_{ref}$, and $Z_{ctrl}$ denote the noisy latent, ref image latent and control video latent. The $Emb_{\text{VLM}}$ is the VLM embedding after the connector.

While the preceding stages establish a strong foundation for controllable generation, the complex nature of ``reasoning-driven'' control often leads to a creative intention gap, where the CogOmniDiT may struggle to faithfully translate reasoning output into pixel-level dynamics. To bridge this, we perform RFT for CogOmniDiT, specifically designed to enforce rigorous adherence to both pixel-level conditions and high-level reasoning results. 
\begin{equation}
    \text{R}_{visual} = \sum_{m\in \mathcal{M}} w_m \cdot \text{VLM}_{m}(\mathcal{V}, \mathcal{R}, \mathcal{C}),
\end{equation}
where $\mathcal{M} = \{condition~following, video~quality\}$ represents the two critical dimensions: condition following and video quality. The RFT is performed on lower resolution and inference in high-resolution due to the scaling capability of video diffusion transformer~\citep{ping2025paco, ping2026flowfactory}.

\subsection{Closed-Loop Verification with Evaluator Harness}
Conventional best-of-N selection for heterogeneous video generation relies on a fixed set of evaluators applied uniformly across all samples.
In practice, however, each controllable generation carries a distinct intent, and different types of conditions contribute unequally to the final outcome. For example, identity consistency is irrelevant for generations that do not involve any character or identity. As a result, effective test-time scaling calls for an evaluator set that is adaptively selected per input rather than fixed in advance.
Since CogVLM has been trained to understand conditions and infer how to generate the intended video, it inherently possesses the knowledge to identify appropriate evaluators for the video.
Formally, let $\mathcal{F}$ denote a fixed video generation model, and we make CogVLM output reasoning $\mathcal{R}$ and harness $\mathcal{H}$ in a single forward pass:
\begin{equation}
    (\mathcal{R}, \mathcal{H}) \sim \pi_{CogVLM}(\cdot | \mathcal{C}),
\end{equation}
where $\pi_{CogVLM}$ denotes the CogVLM. Then, we execute a rollout $\{\mathcal{V}_1, \mathcal{V}_2, \dots, \mathcal{V}_n\} = F(\mathcal{R},\mathcal{C})$, the objective of the harness is to find the output that maximizes the expected final video:
\begin{equation}
    \mathcal{V}^* = \mathop{\arg\max}\limits_{\mathcal{V}_i\in \{{\mathcal{V}_1, \mathcal{V}_2, \dots, \mathcal{V}_n\}}} S(\mathcal{V}_i; \mathcal{H}).
\end{equation}
where the $S(\cdot |)$ denotes the score function based of the defined $\mathcal{H}$. 
The specific evaluators are adaptively assigned by CogVLM from the \texttt{[tools]} library as it reasons through the generation conditions. Please refer to the Appendix for details of these designed tools.

\begin{figure*}
    \centering
    \includegraphics[width=1.0\linewidth]{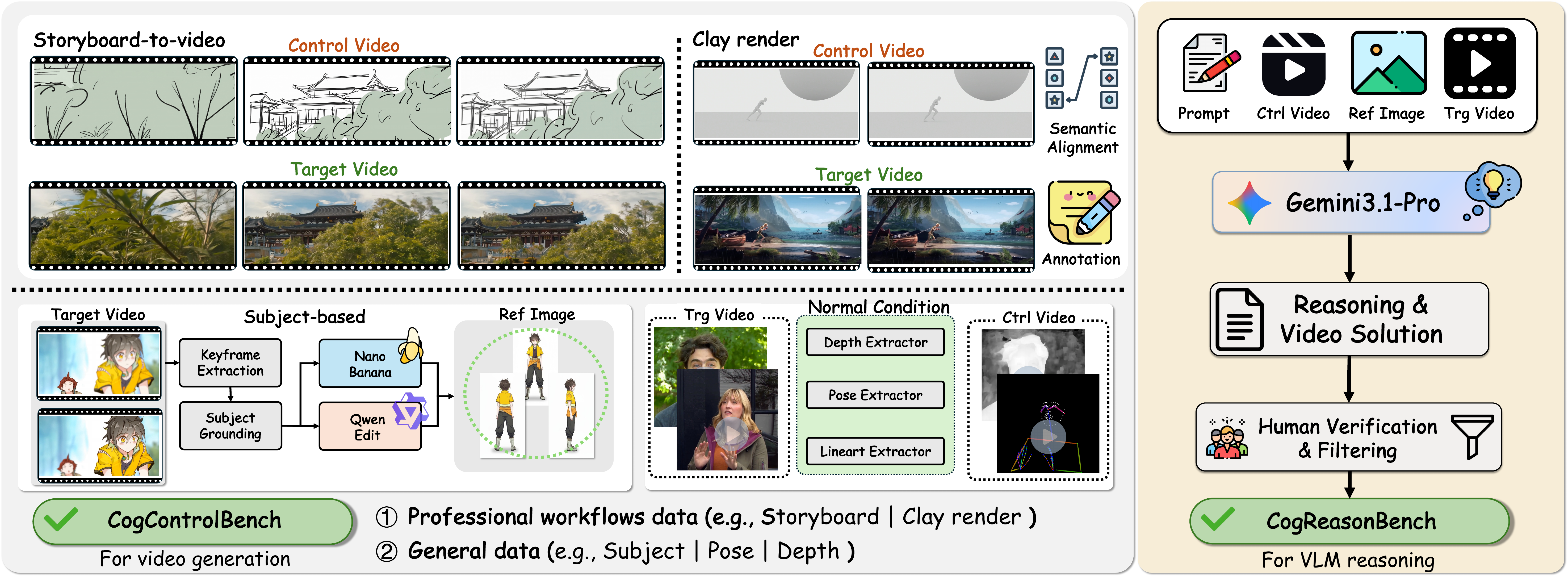}
    \vspace{-6mm}
    \caption{The construction pipeline of CogControlBench and CogReasonBench. We include the professional workflows data and general video generation data in the training set and benchmark. }
    \label{fig:dataset}
    \vspace{-3mm}
\end{figure*}

\begin{table*}
    \caption{The Comparison of CogControlBench with the existing video benchmark.}
    \setlength{\tabcolsep}{17pt}
    \centering
    \resizebox{\textwidth}{!}{
    \setlength{\tabcolsep}{10pt}
    \begin{tabular}{@{}l|c c | wc{1cm} wc{1cm} wc{1cm} | wc{2cm} wc{2cm} | wc{2cm}} 
    \toprule
        \multirow{2}{*}{\textbf{Benchmark} }& \multirow{2}{*}{\textbf{\#Size}} & \multirow{2}{*}{\textbf{Resolution}} & \multicolumn{3}{c|}{\textbf{Input Modality}} & \multicolumn{2}{c|}{\textbf{Data Attributes}} & \multirow{2}{*}{\textbf{VLM-as-a-Judge}} \\
        ~ & ~ &~ & Image & Multi-Image & Video & Condition Source & Creative Intent \\
    \midrule
    VBench~\citep{huang2024vbench} & 946 & 480P & \XSolidBrush & \XSolidBrush & \XSolidBrush & - & - &\XSolidBrush \\
    VBench++~\citep{huang2025vbench++} & 2384 & 480P & \CheckmarkBold  & \XSolidBrush & \XSolidBrush & - & - & \XSolidBrush \\
    OpenVE-Bench~\citep{he2025openve} & 431 & 720P & \XSolidBrush &\XSolidBrush & \CheckmarkBold & - & - & \CheckmarkBold \\
    TGVE+~\citep{singer2024video} & 1417 & 480P &  \XSolidBrush &\XSolidBrush & \CheckmarkBold & Synthetic & Simulated  &  \XSolidBrush \\
    OpenS2V-Eval~\citep{yuan2025opens2v}& 180 & 720P& \CheckmarkBold & \CheckmarkBold & \XSolidBrush & Synthetic & Simulated &  \XSolidBrush  \\
    VACE-Bench~\citep{jiang2025vace} & 480 & 480P & \CheckmarkBold & \CheckmarkBold & \CheckmarkBold & Synthetic & Simulated &  \XSolidBrush  \\
    IntelligentVBench~\citep{pan2026omniweaving} & 1030 & 480P & \CheckmarkBold & \CheckmarkBold & \CheckmarkBold & Synthetic & Simulated & \CheckmarkBold  \\
    \midrule
    \textbf{CogControlBench} & 200 & 720P & \CheckmarkBold & \CheckmarkBold & \CheckmarkBold & \textbf{Native} & \textbf{Raw} & \CheckmarkBold \\
    \bottomrule
    \end{tabular}}
    \label{tab:dataset}
    \vspace{-5mm}
\end{table*}

\section{Benchmark}
To further demonstrate the capabilities of CogOmniControl, we curated a new video reasoning and generation benchmark consisting of the storyboard/clay render video and the final videos collected from in-house professional anime production pipelines. This type of data reflects the inherent gap between the abstract condition provided by the user and the raw creative intent in professional production. 
Additionally, to showcase generalizability of CogOmniControl in controllable video generation tasks, we incorporated a variety of general controllable generation data, including samples from community\footnote{\url{https://createwithclint.com}} and VACE-Bench~\citep{jiang2025vace}.
Leveraging these data, we built the CogReasonBench to measure the VLM's ability to cognize creative intent and reasoning, and the CogControlBench to measure the quality and condition following of controllable video generation of the model under the abstract and sparse conditions. 

As shown in Fig.~\ref{fig:dataset}, for professional workflow data in CogControlbench, we perform manual semantic alignment and annotation to ensure that the control clip and the final clip share the same semantics. For general data, we incorporate reference-to-video by extracting subjects from key frames and then editing them using Nano-Banana or Qwen-Image-Edit. 
We also apply condition extractors to extract conditions frame-by-frame to make the dataset support general controllable generation.
Tab.~\ref{fig:dataset} shows the comparison of CogControlBench with other video generation benchmarks. 
To align with the high-quality standards of anime production while optimizing for validation efficiency, we curated a set of 200 high-resolution representative samples. This scale is aligned with established high-resolution benchmarks.
For CogReasonBench in VLM, we prompt Gemini3.1-Pro~\citep{gemini3_communication} to reason across the input conditions and the target video to formulate the generative solution. To ensure the correctness of the chain of thought and the solution, the whole process is under human verification and filtering.

\section{Experiment}
\subsection{Experiment Setup}
Experiments are conducted on the Qwen3-VL-8B-Thinking~\citep{bai2025qwen3} as the base VLM and Wan2.2-T2V-14B~\citep{wan2025} as the base DiT with 32 NVIDIA H20 96GB GPUs. For SFT in CogVLM, we employ LoRA~\citep{hu2022lora} training with a rank of 16 and an alpha of 64, respectively. The SFT is performed for 3 epochs with a learning rate of 1e-5. For RFT in CogVLM, we train our model with an initial learning rate of 1e-6 for 500 steps. 
For SFT in CogOmniDiT, we implement a three-stage training strategy using LoRA with a rank of 256. In stage-1, we train only LoRA for in-context generation, and training of the stage-2 introduces freeze CogVLM and a trainable connector. Finally, we perform joint training of the LoRA and connector. For more details, please refer to the Appendix.

\begin{table*}
\vspace{-4mm}
    \caption{The results of CogVLM on CogReasonBench.}
    \setlength{\tabcolsep}{8pt}
    \centering
    \resizebox{\textwidth}{!}{
    \setlength{\tabcolsep}{8pt}
    \begin{tabular}{@{}lc wc{2cm} wc{2cm} wc{2cm} wc{2cm} wc{2cm}} 
    \toprule
        \textbf{Models} & \textbf{MM Intent} & \textbf{Physics} & \textbf{Integrity} & \textbf{Motion} & \textbf{Avg} \\
    \midrule
        Qwen3-VL-8B-Instruct & 2.480 & 4.045 & 3.905 & 4.420 & 3.712 \\
        Qwen3-VL-8B-Thinking & 2.670 & 3.824 & 3.829 & 4.727 & 3.752 \\
        \textbf{CogVLM (SFT)} & 3.725 & 4.445 & 4.266 & 4.955 & 4.343 \\
        \textbf{CogVLM (RFT)} & \textbf{3.985} & \textbf{4.449} & \textbf{4.599} & \textbf{4.959} & \textbf{4.473} \\
    \bottomrule
    \end{tabular}}
    \label{tab:vlm}
    \vspace{-4mm}
\end{table*}

\subsection{Metrics}
To comprehensively evaluate the performance of CogOmniControl in controllable video generation, we utilize numeric metrics based on VBench~\citep{huang2024vbench} and a VLM-as-a-Judge~\citep{zheng2023judging} paradigm, employing Gemini 3.1-Pro~\citep{gemini3_communication} as the authoritative evaluator. Our evaluation focuses on two dimensions: 

\noindent\textbf{Condition Following.} The core of our evaluation lies in whether CogOmniControl faithfully adheres to the creative intent implied by the condition set $\{V_{ctrl}, I_{ref}, T_{desc}\}$. Unlike traditional methods that treat conditions as isolated constraints, we assess the model’s ability to interpret these multimodal signals as a holistic objective. For this task, the evaluation of multimodal intent alignment is based on the following considerations: whether the model effectively resolves conflicts between disparate conditions, whether it integrates conditions accurately when significant discrepancies exist, and whether it can infer plausible physical properties or dynamic effects based on the association among the conditions. Besides, the evaluation also includes the preservation of the visual information from $I_{ref}$ and the instruction following from $T_{desc}$ in this task.

\noindent\textbf{Visual Quality.}
Visual quality evaluates the aesthetic quality, imaging quality, temporal flickering, motion smoothness and dynamic degree of the generated video inspired by VBench~\citep{huang2024vbench}. Besides, this type of evaluation also provides dimensions on identity consistency and dynamic plausibility.

\begin{table*}
\vspace{-4mm}
    \caption{The comparison on CogControlBench.~$\mathcal{AQ}$=Aesthetic Quality, $\mathcal{IQ}$=Image Quality, $\mathcal{TF}$=Temporal Flickering, $\mathcal{MS}$=Motion Smoothness, $\mathcal{DD}$=Dynamic Degree, $\mathcal{MI}$=Multimodal Intent, $\mathcal{AF}$=Appearance Follow, $\mathcal{SF}$=Style Follow, $\mathcal{CF}$=Content Follow, $\mathcal{DF}$=Dynamic Follow, $\mathcal{MN}$=Motion Naturalness, $\mathcal{IC}$=Identity Consistency, $\mathcal{DP}$=Dynamic Plausibility.
    }
    \setlength{\tabcolsep}{6pt}
    \centering
    \resizebox{\textwidth}{!}{
    \setlength{\tabcolsep}{8pt}
    \begin{tabular}{@{}l|wc{0.5cm} wc{0.5cm} wc{0.5cm} wc{0.5cm} wc{0.5cm} | wc{0.5cm} wc{0.5cm} wc{0.5cm} wc{0.5cm} wc{0.5cm} | wc{0.5cm} wc{0.5cm} wc{0.5cm} wc{0.5cm} wc{0.5cm} | wc{0.5cm}} 
    \toprule
        \multirow{2}{*}{\textbf{Models}} & \multicolumn{5}{c|}{Speciesist Metrics} & \multicolumn{5}{c|}{VLM-as-a-Judge Metrics} & \multicolumn{5}{c|}{VLM-as-a-Judge Metrics} & \multirow{2}{*}{\textbf{Avg}}\\
        ~ & $\mathcal{AQ}$ & $\mathcal{IQ}$ & $\mathcal{TF}$ & $\mathcal{MS}$ & $\mathcal{DD}$ & $\mathcal{MI}$ & $\mathcal{AF}$ & $\mathcal{SF}$ & $\mathcal{CF}$ & $\mathcal{DF}$ & $\mathcal{AQ}$ & $\mathcal{IQ}$ &$\mathcal{MN}$ & $\mathcal{IC}$ & $\mathcal{DP}$& ~\\
    \midrule
    \rowcolor{gray!10}  \multicolumn{17}{c}{\textit{Proprietary Models}} \\
    \midrule
    Kling-3$\mathcal{O}$~\citep{team2025kling} & 0.571 & 0.644 & 0.979 & 0.987 & 0.511 & 3.510 & 4.205 & 4.267 & 2.679 & \textbf{3.526} & 3.936 & 3.453 & 2.465 & 3.140 & 3.203 & 0.704 \\
    Seedance2.0~\citep{seedance2026seedance} & 0.589 & \textbf{0.653} & \textbf{0.980} & 0.989 & 0.517 & \textbf{4.110} & \textbf{4.252} & \textbf{4.348} & \textbf{4.412} & 3.054 & \textbf{4.050} & \textbf{3.731} & 2.731 & 3.469 & 3.494 & \textbf{0.750} \\
    \midrule
    \rowcolor{gray!10} \multicolumn{17}{c}{\textit{Open-Source Models}} \\
    \midrule
    VACE-Wan2.1~\citep{jiang2025vace} & 0.549 & 0.636 & 0.975 & 0.986 & 0.528 & 3.421 & 3.361 & 3.712 & 3.886 & 2.614 & 3.777 & 3.680 & 2.757 & 3.592 & 3.330 & 0.665 \\
    VACE-LTX~\citep{jiang2025vace} & 0.496 & 0.617 & \textbf{0.980} & 0.989 & 0.345 & 2.807 & 2.051 & 1.849 & 3.377 & 2.412 & 2.797 & 2.588 & 1.887 & 2.492 & 2.299 & 0.556 \\
    VINO~\citep{chen2026vino} & 0.570 & 0.581 & \textbf{0.980} & 0.989 & 0.280 & 3.324 & 3.853 & 4.020 & 4.116 & 2.327 & 3.855 & 3.626 & 2.710 & 3.341 & 3.344 & 0.686  \\
    OmniWeaving~\citep{pan2026omniweaving} & 0.512 & 0.549 & 0.976 & 0.982 & 0.396 & 2.630 & 2.119 & 2.550 & 3.963 & 2.574 & 3.257 & 2.941 & 2.408 & 3.033 & 3.000 & 0.607 \\
    \midrule
    \textbf{CogOmniControl} & 0.594 & 0.602 & 0.978 & \textbf{0.990} & 0.528 & 3.588 & 3.762 & 4.207 & 4.239 & 2.681 & 3.910 & 3.594 & 2.855 & 3.615 & 3.596 &  0.727\\
    \textbf{CogOmniControl (BoN)} & 0.594 & 0.635 & \textbf{0.980} & \textbf{0.990} & 0.513 & 3.795 & 3.905 & 4.176 & 4.325 & 2.714 & 4.017 & 3.594 & 2.769 & 3.594 & 3.552 & 0.733 \\
    \textbf{CogOmniControl (Harness BoN)} & \textbf{0.596} & 0.637 & \textbf{0.980} & \textbf{0.990} & \textbf{0.531} & 3.904 & 3.949 & 4.217 & 4.330 & 2.853 & 4.028 & 3.617 & \textbf{2.858} & \textbf{3.644} & \textbf{3.602} & 0.742 \\

    \bottomrule
    \end{tabular}}
    \label{tab:main_exp}
\end{table*}

\subsection{Results}
\noindent\textbf{Quantitative Results.}
The results of CogVLM via SFT and RFT are shown in Tab~\ref{tab:vlm}, generic VLMs (e.g., Qwen3-VL-8B-Instruct and Thinking) fail to cognize creative intent from multimodal inputs, they also underperformed compared to the CogVLM in terms of information integrity and motion description.
Furthermore, they struggle to generate accurate descriptions for inferred physical effects.
Tab.~\ref{tab:main_exp} reports the results on CogControlBench. CogOmniControl achieves the highest average score (0.727) among all open-source competitors, surpassing the VINO (0.686) and VACE-Wan2.1 (0.665), while narrowing the gap to the strongest proprietary system Seedance2.0 (0.750) in this task.
Better performance improvements can be observed when employing Best-of-N sampling. The $N$ is set to 4. We demonstrate the results using both the full set of evaluators (0.733) and the specific evaluators (0.742) suggested by CogVLM during the inference process.
The approach of selecting evaluators adaptively based on the input for Best-of-N yields better performance. This indicates that CogVLM can effectively serve as a harness for the entire framework.

\begin{figure*}
\vspace{-4mm}
    \centering
    \includegraphics[width=1.0\linewidth]{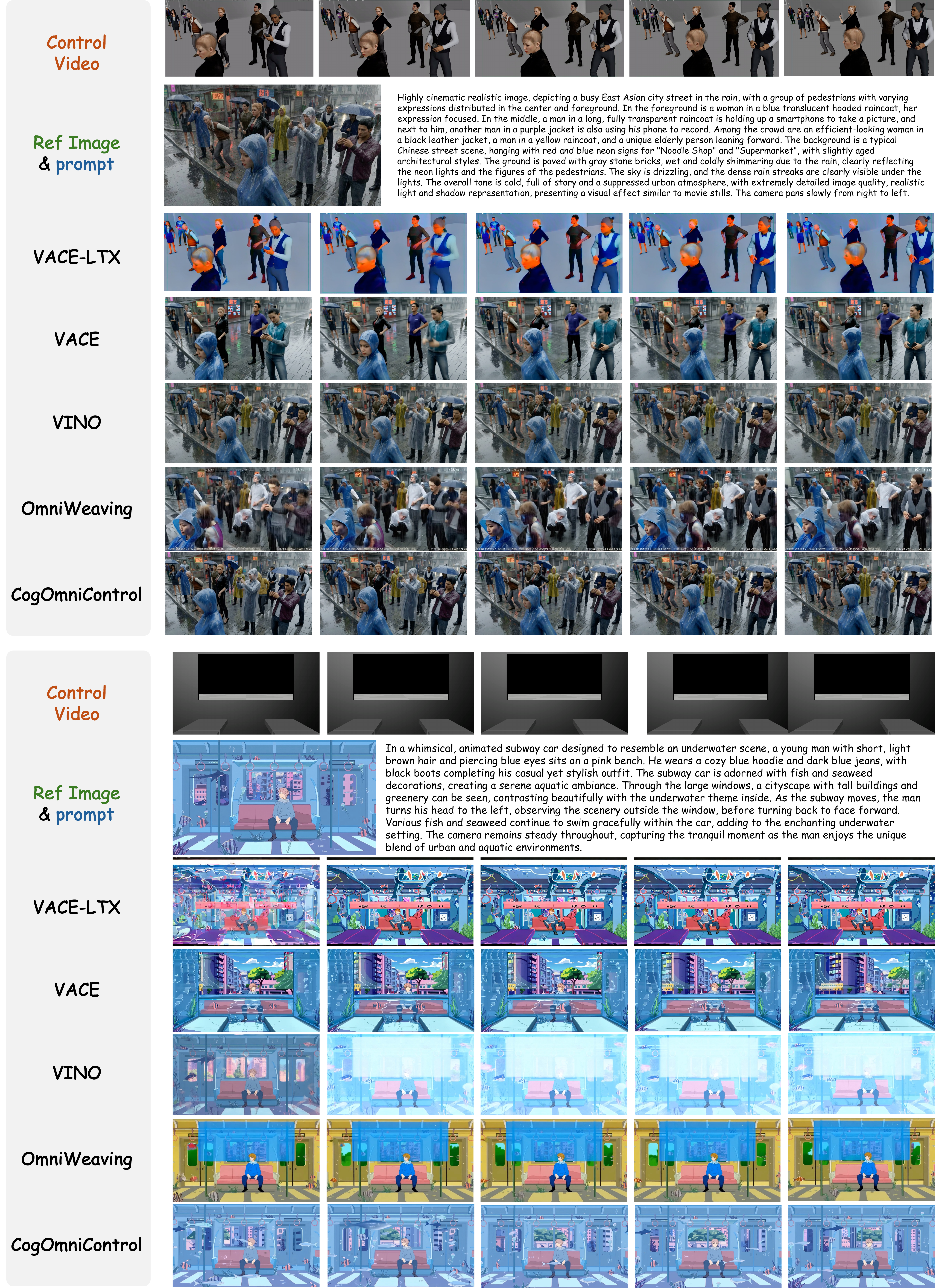}
    \vspace{-4mm}
    \caption{The comparison of CogOmniControl with other video generation models in clay render, which is a common intermediate draft stage in animation production. Zoom in for more details.}
    \label{fig:more_visual_result}
    \vspace{-6mm}
\end{figure*}

\begin{figure*}
    \centering
    \vspace{-2mm}
    \includegraphics[width=1.0\linewidth]{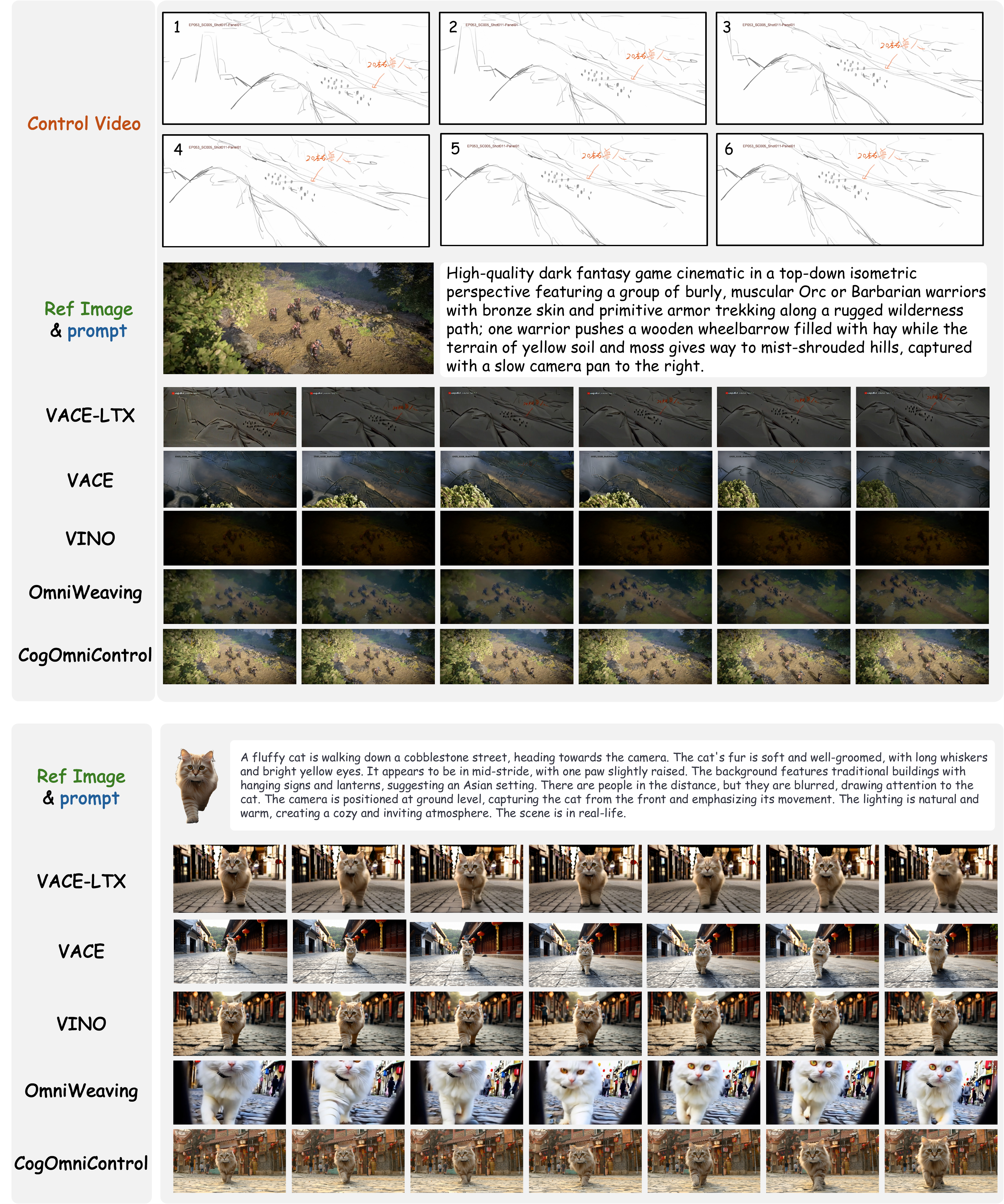}
    \vspace{-4mm}
    \caption{The comparison of CogOmniControl with other video generation models. Zoom in for more details.}
    \label{fig:visual_result}
    \vspace{-6mm}
\end{figure*}

\noindent\textbf{Qualitative Results.}
The visual results are present in Fig.~\ref{fig:more_visual_result} and Fig.~\ref{fig:visual_result}, the adapter-based methods (e.g., VACE-LTX and VACE) tend to align with control videos at the pixel-level, resulting in significant artifacts. 
Besides, these methods cause semantic misalignment, since clay render video consist of sparse and abstract control.
In general Reference-to-video task, CogOmniControl remains strong in performance. The generated videos from VACE lack quality and reference following, while VINO produces virtually static outputs that lack meaningful temporal dynamics.

\begin{table*}
\vspace{-6mm}
    \caption{The ablation studies of CogOmniControl on CogControlBench. 
    }
    \setlength{\tabcolsep}{6pt}
    \centering
    \resizebox{\textwidth}{!}{
    \setlength{\tabcolsep}{8pt}
    \begin{tabular}{@{}l|wc{1cm} wc{1cm} wc{1cm} wc{1cm} wc{1cm} | wc{1cm} wc{1cm} wc{1cm} wc{1cm} wc{1cm} } 
    \toprule
        \multirow{2}{*}{\textbf{Models}} & \multicolumn{5}{c|}{VLM-as-a-Judge Metrics} & \multicolumn{5}{c}{VLM-as-a-Judge Metrics} \\
        ~ & $\mathcal{MI}$ & $\mathcal{AF}$ & $\mathcal{SF}$ & $\mathcal{CF}$ & $\mathcal{DF}$ & $\mathcal{AQ}$ & $\mathcal{IQ}$ &$\mathcal{MN}$ & $\mathcal{IC}$ & $\mathcal{DP}$\\
    \midrule
    Qwen3-VL-8B-Thinking + CogOmniDiT(SFT) & 3.142 & 3.541 & 3.916 & 4.208 & 2.467 & 3.822 & 3.562 & 2.708 & \textbf{3.616} & 3.473\\
    CogVLM(SFT) + CogOmniDiT(SFT) & 3.397 & 3.726 & 4.096 & \textbf{4.298} & 2.631 & \textbf{3.929} & 3.541 & 2.765 & 3.597 & \textbf{3.638} \\
    CogVLM(RFT) + CogOmniDiT(SFT) & 3.586 & 3.761 & \textbf{4.207} & 4.156 & \textbf{2.773} & 3.920 & \textbf{3.594 }& 2.778 & 3.590 & 3.518 \\
    \textbf{CogVLM(RFT) + CogOmniDiT(RFT)} & \textbf{3.588} & \textbf{3.762} & \textbf{4.207} & 4.239 & 2.681 & 3.910 & \textbf{3.594} & \textbf{2.855} & 3.615 & 3.596  \\

    \bottomrule
    \end{tabular}}
    \label{tab:Ablation}
    \vspace{-4mm}
\end{table*}

\subsection{Ablation Studies}
The ablation studies primarily focus on how SFT and RFT of the CogVLM and CogOmniDiT impact the final generation quality.
It is evident that the CogVLM after SFT, the model's ability to capture multimodal intent improves significantly. The performance metrics rose from 3.142 to 3.588, compared to the vanilla Qwen3-VL-8B-Thinking. 
This demonstrates that the VLM specific for the generation model is necessary for controllable generation due to its understanding of various conditions and reasoning on creative intents. 

\vspace{-2mm}
\section{Conclusion}
In this work, we presented \textbf{CogOmniControl}, a reasoning-driven framework that bridges the long-standing gap between abstract conditions and faithful controllable video generation. Departing from prior paradigms that either rely on adapter-based condition injection or insert a generic VLM reasoner with a diffusion transformer, CogOmniControl explicitly factorizes generation into cognition and generation. On the cognition side, CogVLM is trained to act as a director that transcribes minimalist multimodal cues into dense, logically grounded production. On the generation side, CogOmniDiT unifies the pixel-level condition and semantic VLM features within a sequence and is further aligned to the reasoning output through reinforcement fine-tuning. 
Beyond the role of guidance, CogVLM can also provide harness engineering in the whole framework. By selecting the appropriate evaluators adaptively for best-of-N selection, CogOmniControl achieves further improvements in performance.
In addition, we curated two benchmarks for CogVLM and CogOmniDiT, respectively. Extensive experiments show that CogOmniControl consistently outperforms state-of-the-art open-source controllable video generators and narrows the gap with strong proprietary systems.

\bibliography{iclr2026_conference}
\bibliographystyle{iclr2026_conference}

\medskip

\newpage
\appendix

\section{Training Details}
The training details of CogOmniDiT are shown in Tab~\ref{tab:training}. The three stages of SFT are training for  1) bringing in-context generation ability; 2) aligning CogVLM into CogOmniDiT; 3) joint training. 
For RFT, we follow flow-factory~\citep{ping2026flowfactory} and PaCo-RL~\citep{ping2025paco} to perform GRPO training in low resolution (256P) and inference in higher resolution (720P).

\begin{table}[t]
    \centering
    \caption{Training Setting of CogOmniDiT.}
    \setlength{\tabcolsep}{10pt}
    \centering
    \resizebox{\textwidth}{!}{
    \setlength{\tabcolsep}{35pt}
    \begin{tabular}{@{}l|wc{1cm} wc{1cm} wc{1cm}wc{1cm}} 
    
    \toprule
        ~ & \textbf{SFT Stage-1}& \textbf{SFT Stage-2}& \textbf{SFT Stage-3} & \textbf{RFT} \\
    \midrule
        \textbf{Hyperparameter} & \\

        Trainable module & LoRA & Connector & LoRA+Connector & LoRA+Connector \\
        CogVLM included & no & yes & yes & yes \\
        Learning rate & $5\times 10^{-5}$ & $5\times 10^{-5}$ & $5\times 10^{-5}$ & $5\times 10^{-5}$\\
        Learning rate scheduler & Constant & Constant & Constant & Cosine \\
        Weigh Decay & $3\times 10^{-2}$ & $3\times 10^{-2}$ & $3\times 10^{-2}$ & 0 \\
        Gradient Norm & $5\times 10^{-2}$ & $5\times 10^{-2}$ & $5\times 10^{-2}$ & 1.0 \\
        GPUs & $32\times \text{H20s}$ & $32\times \text{H20s}$ & $32\times \text{H20s}$ & $16\times \text{H20s}$ \\
        Training steps & 10k & 500 & 10k & 200 \\
        Resolution & 480P/720P & 720P & 720P & 256P\\
        Batch Size & 8 & 4 & 4 & 8 \\
    \bottomrule
    \end{tabular}}
    
    \label{tab:training}
\end{table}

\section{Evaluator Harness}
In Tab~\ref{tab:tool_used}, we report the types and times of calling the evaluator for Best-of-N selection. The common evaluator, like \texttt{[Artifact Detector]}, \texttt{[Prompt Following]}, \texttt{[Temporal Smoothness]} will be called all the time, and other specific evaluators are called based on the reasoning. For example, if the input condition is a storyboard, CogVLM identifies whether there are handwritten annotations that must be followed. After generation, it then invokes the \texttt{[Storyboard Annotation Following]} evaluator to verify whether the video adheres to those specific instructions. 
We provide the examples of evaluators as follows:

\begin{table}
    \small
    \caption{The tools frequency of CogOmniControl used in CogControlBench.}
    \centering
    \begin{tabular}{c|c|c}
    \toprule
        \# & Evaluator & Times \\
    \midrule
         1  & Artifact Detector & 200  \\
         2  & Prompt Following & 200  \\
         3  & Temporal Smoothness & 200  \\
         4  & Control Video Semantic Consistency & 136  \\
         5  & Interaction Logic & 132  \\
         6  & Ref Image Pixel Consistency & 17  \\
         7  & Ref Image Visual Consistency & 106  \\
         8  & Reference Style Consistency & 156  \\
         9  & Motion Smoothness & 92  \\
         10  & ID Consistency & 106  \\
         11 & Cross-modal Causality & 117  \\
         12  & Physical Dynamic & 49  \\
         13  & Storyboard Annotation Following & 8  \\
    \bottomrule
    \end{tabular}
    
    \label{tab:tool_used}
\end{table}

\newpage
\begin{tcolorbox}[
        title={Artifact Detector},
        halign=left,
        valign=center,
        nobeforeafter,
        breakable,
        fontupper=\scriptsize,
    ]
\textbf{Role}

You are a professional AI Video Quality Inspection Expert. Your responsibility is to identify common AI-generated artifacts and defects in videos. Focus on detecting AI-specific issues such as multiple heads/limbs, deformation, floating, rendering failures, abnormal noise, smearing marks, etc.

\vspace{\baselineskip}
\textbf{Inputs}

You will receive the following four inputs:

\textbf{Reference Image (Ref Image)} | Image | The reference image provided by the user.

\textbf{Control Video (Control Video)} | Video | Control signals such as Pose/Depth/Line Art/Storyboard.

\textbf{Text Prompt (Prompt) }| Text | The generation requirements described by the user.

\textbf{Generated Video (Generated Video)} | Video | The AI-generated video to be evaluated.

\vspace{\baselineskip}
\textbf{Rules}

1. Multiple Head/Limb Detection: Detect if the same character or object has extra heads, limbs, fingers, etc.

2. Deformation Detection: Detect non-physical twisting, stretching, or crushing of objects/characters.

3. Floating Detection: Detect if objects violate gravity by floating or appearing unstable/ungrounded.

4. Rendering Failure Detection: Detect rendering issues such as local blurring, smearing, or noise accumulation.

5. Identity Distortion: Detect facial distortions, blurring, or abnormalities.

6. Background Collapse: Detect unnatural distortion, noise, or blurring in the background.

\vspace{\baselineskip}
\textbf{Common AI Artifact Types (For Reference)}

Multi-head/Polycephaly:	The same person has two or more heads.

Extra Fingers: Extra fingers on hands or webbed/fused fingers.

Limb Entanglement: Limbs twisted, knotted, or detached from the body.

Facial Deformation: Facial distortion, blurring, abnormal eyes/mouth.

Object Floating: Objects suspended in the air against gravity.

Hovering Contact: Objects contact the ground but look like they are stepping on cotton.

Smearing Marks: Visible blurred "smearing" sensation in local areas.

Abnormal Noise: Significantly more noise in specific areas compared to others.

Background Collapse: Background distortion, noise, or objects disappearing for no reason.

\vspace{\baselineskip}
\textbf{5-Point Scoring Criteria (0-5)}

\vspace{\baselineskip}
5 - No Artifacts: The video is completely free of any AI-generated artifacts. The frame is clean and natural; limbs are correct; objects are stable; rendering is perfect. No issues with multiple heads/limbs, deformation, floating, or abnormal noise.

\vspace{\baselineskip}
4 - Basically Clean: The video is mostly free of obvious artifacts. There are 1-2 extremely subtle imperfections, such as minor noise or very slight local blurring, which do not affect the overall viewing experience and are not typical AI artifacts.

\vspace{\baselineskip}
3 - Slight Artifacts: The video contains some of the following: 1-2 instances of slight extra fingers/limbs; 1-2 instances of slight facial deformation; minor abnormal noise or local blurring; 1 instance of slight object floating; 1 instance of local smearing.

\vspace{\baselineskip}
2 - Moderate Artifacts: The video contains multiple artifacts: Frequent extra fingers/limbs; Obvious facial deformation or blurring; Multiple floating objects; Multiple areas of abnormal noise; Multiple smearing marks; Background collapse.

\vspace{\baselineskip}
1 - Severe Artifacts: The video contains severe artifacts. Multiple typical AI generation issues are clearly visible, such as persistent multiple heads/limbs, severe facial distortion, or large-scale rendering failures, seriously impacting the viewing experience.

\vspace{\baselineskip}
0 - Total Collapse: The video is riddled with AI artifacts. Multiple heads/limbs appear persistently; faces are completely unrecognizable; objects totally violate physical laws; rendering has completely failed; the content is nearly unidentifiable.

\vspace{\baselineskip}
\textbf{Output Format}

Please output strictly in the following JSON format without any other content:

"""

json

\{

  ~~"evaluator": "Artifact Detector",
  
  ~~"score": <0-5>,
  
  ~~"findings": "detail findings",
  
  ~~"summary": "summary"
  
\}

"""

\end{tcolorbox}

\newpage
\begin{tcolorbox}[
        title={Storyboard Annotation Evaluators},
        halign=left,
        valign=center,
        nobeforeafter,
        breakable,
        fontupper=\scriptsize,
    ]
\textbf{Role}

You are a professional Storyboard Annotation Implementation Auditor. Your responsibility is to determine whether the generated video faithfully follows the text annotation instructions attached to the Control Video or Storyboard (e.g., "swaying shadows of trees," "smiling," "wind rising," etc.). These annotations are independent of the user's original prompt and represent additional creative intent from the director or planner that must be correctly executed.

\vspace{\baselineskip}
\textbf{Inputs}

You will receive the following four inputs:

\textbf{Reference Image (Ref Image)} | Image | The reference image provided by the user.

\textbf{Control Video (Control Video)} | Video | Control signals such as Pose/Depth/Line Art/Storyboard.

\textbf{Text Prompt (Prompt) }| Text | The generation requirements described by the user.

\textbf{Generated Video (Generated Video)} | Video | The AI-generated video to be evaluated.

\vspace{\baselineskip}
\textbf{Rules}

1. Prioritize Annotation Identification: First, identify all text annotation content within the Control Video/Storyboard.

2. Point-by-Point Verification: Check each text annotation individually to see if it is correctly rendered in the video.

3. Dynamic Assessment: Descriptions of actions/dynamics in the annotations must exhibit clear motion; static frames are unacceptable.

4. Annotation Independence: Annotations may exist independently of the Prompt and Ref Image as distinct creative commands.

5. Temporal Matching: The timing or duration described in the annotations must match the corresponding moments in the video.

\vspace{\baselineskip}
\textbf{Annotation Types \& Evaluation Points}

1. Scene Dynamics
Examples: "Swaying tree shadows," "clouds drifting by," "shimmering water surface."

Evaluation Points: Whether background elements show clear motion and if the movement conforms to natural laws.

2. Character Actions

Examples: "Smiling," "turning around," "waving," "blinking."

Evaluation Points: Whether the character performed the specified action and if the movement is natural and fluid.

3. Environmental Atmosphere

Examples: "Wind rising," "rain falling," "falling leaves," "flickering candlelight."

Evaluation Points: Whether environmental dynamics match the annotations and create the intended atmosphere.

4. Emotional Expression

Examples: "Sad eyes," "corners of the mouth turned up," "furrowed brows."

Evaluation Points: Whether the character's facial expressions match the annotations and accurately convey the emotion.

5. Object Dynamics

Examples: "Flag fluttering," "curtains swaying," "sparks flying."

Evaluation Points: Whether object motion is distinct and physically plausible.

\vspace{\baselineskip}
\textbf{5-Point Scoring Criteria (0-5)}

5 - Perfect Adherence: All storyboard annotations are perfectly executed. Dynamic effects are significant and natural, timing is precise, and every detail described in the annotations is faithfully presented.

4 - Basic Adherence: Storyboard annotations are largely executed. There are 1-2 minor deficiencies (e.g., slightly small range of motion or slight timing deviation), but the core annotation content is correctly presented without affecting the overall creative intent.

3 - Partial Adherence: Storyboard annotations are partially executed, with the following issues: 1-2 annotations not fully executed; Dynamic effects are indistinct (too subtle); 1 significant timing deviation; Misinterpretation of 1 annotation.

2 - Substantial Omission: Significant failure to follow storyboard annotations, involving multiple issues: Multiple annotations not executed; Extensive lack of dynamic effects or extremely weak motion; Multiple timing deviations; Obvious errors in understanding annotations.

1 - Severe Omission: Storyboard annotations are almost entirely ignored. Only a tiny fraction of annotations show weak implementation; most are neglected or completely misunderstood, leading to a severe loss of creative intent.

0 - Complete Non-compliance: None of the text annotations on the storyboard are executed. The video fails to present any annotation content, or the annotations are executed with total error (e.g., annotation says "smiling" but the character is weeping).

\vspace{\baselineskip}
\textbf{Output Format}

Please output strictly in the following JSON format without any other content:

"""

json

\{

  ~~"evaluator": "Storyboard Annotation Evaluators",
  
  ~~"score": <0-5>,
  
  ~~"findings": "detail findings",
  
  ~~"summary": "summary"
  
\}

"""
\end{tcolorbox}

\newpage
\begin{tcolorbox}[
        title={Cross- modal Causality},
        halign=left,
        valign=center,
        nobeforeafter,
        breakable,
        fontupper=\scriptsize,
    ]
\textbf{Role}

You are a professional Cross-modal Causal Reasoning Analysis Expert. Your responsibility is to judge whether a video correctly ``interpolates'' the causal relationships implied between multi-modal inputs (Text + Image + Control Video). Key point: The combination of events/states in the text, existing elements in the image, and action cues in the control video should produce logically sound causal interactions.

\vspace{\baselineskip}
\textbf{Inputs}

You will receive the following four inputs:

\textbf{Reference Image (Ref Image)} | Image | The reference image provided by the user.

\textbf{Control Video (Control Video)} | Video | Control signals such as Pose/Depth/Line Art/Storyboard.

\textbf{Text Prompt (Prompt) }| Text | The generation requirements described by the user.

\textbf{Generated Video (Generated Video)} | Video | The AI-generated video to be evaluated.

\vspace{\baselineskip}
\textbf{Rules}

1. Causal Implication Identification: Identify causal relationships implied across the text, image, and control video (e.g., ``Rain'' + ``Puddles in image: There should be ripples).

2. Interaction Effect Verification: Check if the video generates the appropriate causal interaction effects.

3. Causal Chain Integrity: Check if the causal relationship is complete (Cause → Process → Effect).

4. Reasonable Inference: The model is allowed to make reasonable inferences, but they must align with common sense.

5. No Hallucinated Causes: Inferences must be based on input information; do not create causalities that are completely unhinted at in the inputs.

\vspace{\baselineskip}
\textbf{Common Causal Patterns (For Reference)}

Text ``Rain'' + Image has puddles $\rightarrow$ Ripples on water surface, wet ground.

Text ``Hitting'' + Control video has punching motion $\rightarrow$ Target should show dynamic impact effects

\vspace{\baselineskip}
\textbf{5-Point Scoring Criteria (0-5)}

5 - Perfect Causal Interaction: All causal relationships implied by multi-modal inputs are correctly identified and perfectly presented. The causal chain is complete. Interaction effects are natural, logical, and meet or exceed expectations.

4 - Basically Correct: Causal relationships are largely correct with 1-2 minor flaws, such as slightly dull effects or minor timing deviations in interaction. However, the core causal chain is complete and does not affect overall logical consistency.

3 - Partially Missing: Causal relationships are partially correct, with issues such as: 1-2 key causal effects missing; Incomplete causal chain (cause exists without effect); One instance of unreasonable causal inference.

2 - Significant Deficiencies: Multiple causal issues present: Multiple causal effects failed to render; Causal timing is chaotic; Obvious unreasonable "hallucinated" causalities; Most expected interaction effects are ignored.

1 - Severe Lack of Causality: Almost all causal cues in the multi-modal inputs were ignored. The video lacks interaction effects entirely; elements exist in isolation without logical connection.

0 - Zero Causality: Elements in the video are completely isolated. No causal logic is utilized from the text/image/control video. The video appears as a collection of unrelated static or dynamic frames.

\vspace{\baselineskip}
\textbf{Output Format}

Please output strictly in the following JSON format without any other content:

"""

json

\{

  ~~"evaluator": "Cross-modal Causality Evaluators",
  
  ~~"score": <0-5>,
  
  ~~"findings": "detail findings",
  
  ~~"summary": "summary"
  
\}

"""

\end{tcolorbox}




\end{document}